\documentclass[journal]{IEEEtran}

\ifCLASSINFOpdf
\else
   \usepackage[dvips]{graphicx}
\fi
\usepackage{url}

\usepackage{bookmark}
\usepackage{graphicx}
\usepackage{multirow}
\usepackage{amsmath,amssymb,booktabs}
\usepackage{color}
\usepackage{stfloats}
\usepackage{hyperref}
\usepackage{etoolbox}
\makeatletter
\patchcmd{\@makecaption}
  {\scshape}
  {}
  {}
  {}
\makeatother
\begin{document}

\title{LFSamba: Marry SAM with Mamba for Light Field Salient Object Detection}

\author{Zhengyi Liu*, Longzhen Wang, Xianyong Fang, Zhengzheng Tu, Linbo Wang
\thanks{Zhengyi Liu, Longzhen Wang, Xianyong Fang, Zhengzheng Tu, and Linbo Wang are with the School of Computer Science and Technology, Anhui University, Hefei, China (e-mail: liuzywen@ahu.edu.cn, 1774537072@qq.com, fangxianyong@ahu.edu.cn, zhengzhengahu@163.com, wanglb@ahu.edu.cn). Zhengyi Liu is the corresponding author. This work is supported by National Natural Science Foundation of China under Grant 62376005.}}

\markboth{Journal of \LaTeX\ Class Files, Vol. 14, No. 8, September 2024}
{Shell \MakeLowercase{\textit{et al.}}: Bare Demo of IEEEtran.cls for IEEE Journals}
\maketitle

\begin{abstract}
A light field camera can reconstruct 3D scenes using captured multi-focus images that contain rich spatial geometric information, enhancing applications in stereoscopic photography, virtual reality, and robotic vision. In this work, a state-of-the-art salient object detection model for multi-focus light field images, called LFSamba, is introduced to emphasize four main insights: (a) Efficient feature extraction, where SAM is used to extract modality-aware discriminative features; (b) Inter-slice relation modeling, leveraging Mamba to capture long-range dependencies across multiple focal slices, thus extracting implicit depth cues; (c) Inter-modal relation modeling, utilizing Mamba to integrate all-focus and multi-focus images, enabling mutual enhancement; (d) Weakly supervised learning capability, developing a scribble annotation dataset from an existing pixel-level mask dataset, establishing the first scribble-supervised baseline for light field salient object detection. \href{https://github.com/liuzywen/LFScribble}{https://github.com/liuzywen/LFScribble}

\end{abstract}

\begin{IEEEkeywords}
SAM, Mamba, multi-focus, light field, salient object detection
\end{IEEEkeywords}

\IEEEpeerreviewmaketitle

\section{Introduction}
\begin{figure}[!htp]
  \centering
  \includegraphics[width=1\linewidth]{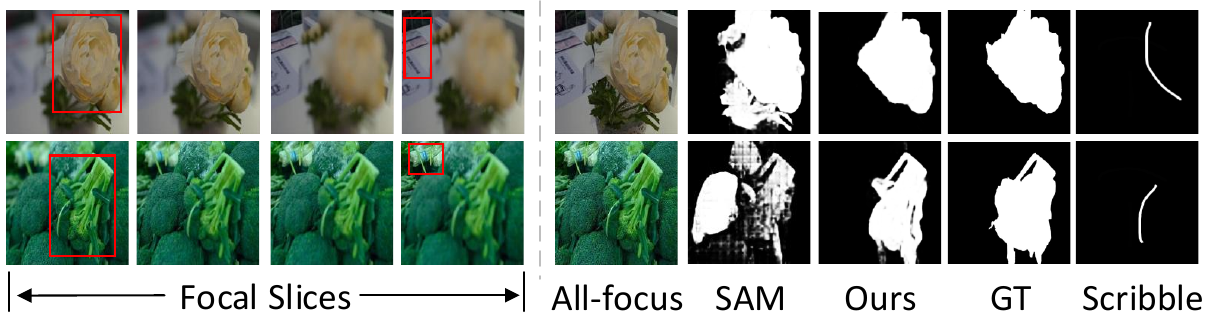}
  \caption{Examples of multi-focus  light field images. Each example consists of a set of focal slices and an all-focus image. Red boxes indicate the focused region. Compared with ``SAM" finetuned exclusively on the all-focus image, ``Ours" finetuned on multi-focus images are closer to the pixel-level mask annotation ``GT". ``Scribble" refers to the sparse annotation.
  \label{fig:example}}
\end{figure}

Light field (LF) cameras \cite{he2023light} play an important role in stereoscopic photography, virtual reality, and robotic vision applications, because they can reconstruct 3D scenes via multi-view and multi-focus images.
%They show the geometry of the observed scene by capturing the color intensity of each pixel and the directions of all incoming light rays.
Multi-view images \cite{yuan2024parallax,zhang2024light} reflect the panoramic views of objects, effectively addressing the occlusion problem, while multi-focus images \cite{wang2020three} perceive the spatial context of objects, benefiting the partition of foreground objects and background. Fig \ref{fig:example} presents examples of multi-focus images, which typically consist of some focal slices  focused at different depth levels and an all-focus image synthesized from all focal slices via photo-montage technique. %\cite{agarwala2004interactive}
Each focal slice asynchronously focuses on different depth positions and blur the others, while the all-focus image simultaneously depicts the full scene appearance and ignore the depth of field of objects.

Salient object detection (SOD) in multi-focus images can extract attractive objects for the observers based on both all focal slices and all-focus images.
Compared with the current predominant method which  finetunes Segment Anything Model (SAM) \cite{kirillov2023segment} using adapters \cite{chen2022adaptformer} exclusively on all-focus images,  our method locates the correct salient objects by perceiving the depth information embedded in multi-focus images, as shown in Fig \ref{fig:example}. Three key perspectives which benefit SOD task for multi-focus LF images will be elaborated.

%In the paper, we focus on three important perspectives which benefit salient object detection task for multi-focus light field images.

%attributing to our proposed feature extraction and fusion method. Moreover, we exploit scribble supervised solution for multi-focus images for the first time and construct the first scribble dataset.

\textit{Feature extraction ability:} Detecting salient objects that attract observers requires a highly effective feature extractor. In the past few years, transformer encoders have outperforms convolutional ones \cite{dosovitskiy2021an}.
%, e.g., VGG \cite{simonyan2014very} and ResNet \cite{he2016deep}.
Based on long range dependency of transformer framework and massive amount of training samples, SAM has been widely used to encode features, showing the discriminative advantage. However, in LF SOD task, it is necessary to encode multiple focal slices along with an all-focus image. To both mitigate the computation cost and enhance feature discrimination, a frozen SAM encoder with a group of fine-tuned adapters is used to encode focal slice features and all-focus features.
%Inspired by  parameter-efficient fine-tuning \cite{fu2023effectiveness},

\textit{Fusion ability:} Fusion in light field includes fusion among different focal slices and fusion between focal slices and all-focus images. The former is the sequence fusion of the images with the same modality, while the later is multi-modal fusion.
Because Mamba can model long-range dependencies in long sequence data \cite{gu2023mamba}, it is adopted to process the sequence of focal slices, extracting implicit spatial structure information from the scenes.
Furthermore, an inter-modal Mamba is designed to achieve multi-modal fusion, highlighting commonalities and suppressing redundancies.

\textit{Weakly supervised learning ability:} Annotation is  important for deep learning model to learn latent mapping from input to output.
Existing methods trained LF SOD models via dense annotation with high labor costs. To eliminate the labelling overhead, a scribble annotation dataset  and a weak supervised learning method is constructed and exploited, respectively. The last column of Fig \ref{fig:example} gives annotation examples which use sparse scribbles to indicate the foreground \cite{li2023hybridvps}.

\section{Method}
\subsection{Model pipeline}
\begin{figure}[!htp]
\center
  \includegraphics[width=1\linewidth]{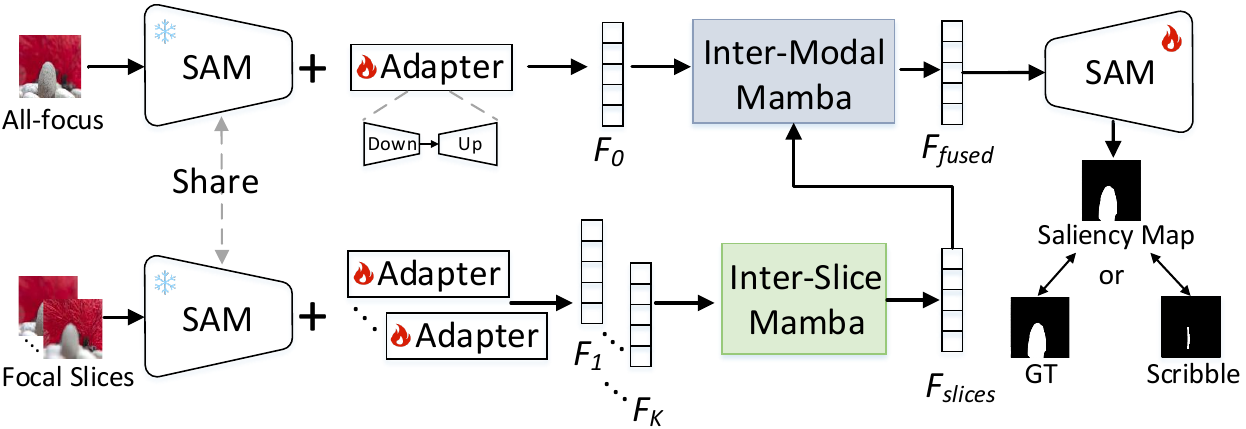}
  \caption{The pipeline of LFSamba.}
  \label{fig:main}
\end{figure}

The proposed LFSamba is a two-stream encoder-decoder framework based on SAM.
In the encoder part,  a shared and frozen SAM encoder with the finetuned adapters is used to extract focal slice features and all-focus features, respectively.
An Inter-Slice Mamba is proposed to integrate all focal slice features in different depth levels. Furthermore, an inter-modal mamba is proposed to fuse focal slice features and all-focus features which are different in modality.
Finally, the fused features are fed into SAM decoder to output the saliency map.
The fully-supervised model is trained with the supervision of pixel-level ground truth, while the weakly-supervised model is trained with the supervision of our constructed scribble annotation.
The whole architecture is shown in Fig. \ref{fig:main}.
\subsection{Feature extraction by large model SAM}
\textit{\textbf{Motivation. }}Recently the large model SAM has been proposed to achieve class-agnostic segmentation task. However, the zero-shot capability of SAM is not sufficiently robust to adapt to LF SOD task. Therefore, fine-tuning SAM using adapters offers an alternative approach for extracting features from light field images.

\textit{\textbf{Design. }}Given an all-focus image $I_r$ and its corresponding focal stack  with $K$ focal slices $\{I_{f_k}\}_{k=1}^K$,
an SAM encoder with $K$+$1$ groups of finetuned adapters is used to generate all-focus features $F_0$ and focal slice features $\{F_k\}_{k=1}^K$.
\begin{small}\begin{equation}
    F_0=\mathcal{E}_{Adapter_0}(I_r)
\end{equation}\end{small}
\vspace{-5mm}
\begin{small}\begin{equation}
   F_k= \mathcal{E}_{Adapter_k}(I_{f_k})(k=1,\cdots,K)
\end{equation}\end{small}%
where $\mathcal{E}_{Adapter_k}(k=0,\cdots,K)$ is a frozen SAM encoder with  the $k$-th group of adapter, subscript $k=0$  is the all-focus indicator, while $k=1,\cdots,K$ are focal slice indicators.
Each group of adapters includes a position adapter and some feature adapters. The position adapter is a max pooling operation followed a convolution operation with a kernel size of 3$\times$3 conducted on position embedding.
It is responsible for adapting the model to the input with a smaller size (256$\times$256) rather than the original size (1024$\times$1024) used in vanilla SAM, aiming to reduce the computation cost.
The feature adapter is a bottleneck structure, which consists of a down-projection layer, a ReLU layer, and an up-projection layer,  in parallel with multi-layer perceptron (MLP) layer of transformer block in vanilla SAM encoder.
It is in charge of extracting modality-aware all-focus feature and focal slice features.
\subsection{Feature integration within slices via inter-slice mamba}
\textit{\textbf{Motivation.}}
The multi-focus image includes a set of focal slices which show the similar appearance and different focused region.
To integrate all the focal slices, many technologies have been applied.
MEANet \cite{jiang2022meanet} conducted concatenation operation in input level, ignoring the correlation among slices.
SA-Net \cite{zhang2021SANet}, DLGLRG \cite{liu2021light}, and LFTransNet \cite{liu2023lftransnet} used 3D convolution, graph network, and transformer, respectively. However, these methods consumed a large computational overhead.
LFNet \cite{zhang2020lfnet}, MoLF \cite{zhang2019memory}, ERNet \cite{piao2020exploit}, and NoiseLF \cite{feng2022learning} employed  ConvLSTM  \cite{shi2015convolutional} to learn spatial structures recurrently, but the local convolutional nature made it less effective at capturing long-range dependencies among  focal slices.

Recently,  Mamba \cite{gu2023mamba}, derived from State Space Models (SSMs) \cite{guefficiently}, has advanced deep learning models due to their capability of modeling long sequences to achieve global receptive fields with linear complexity.
Vision Mamba \cite{zhuvision} integrated SSM with bidirectional scanning, making each patch related to another.
Meanwhile, VMamba \cite{liu2024vmamba} extended scanning in four directions in the proposed 2D Selective Scan (SS2D).
To excavate latent information in the focal slices, SS2D is adopted to integrate all the focal slices.

\begin{figure}[!htp]
\centering
\begin{tabular}{ccc}
\includegraphics[width = 0.8\linewidth]{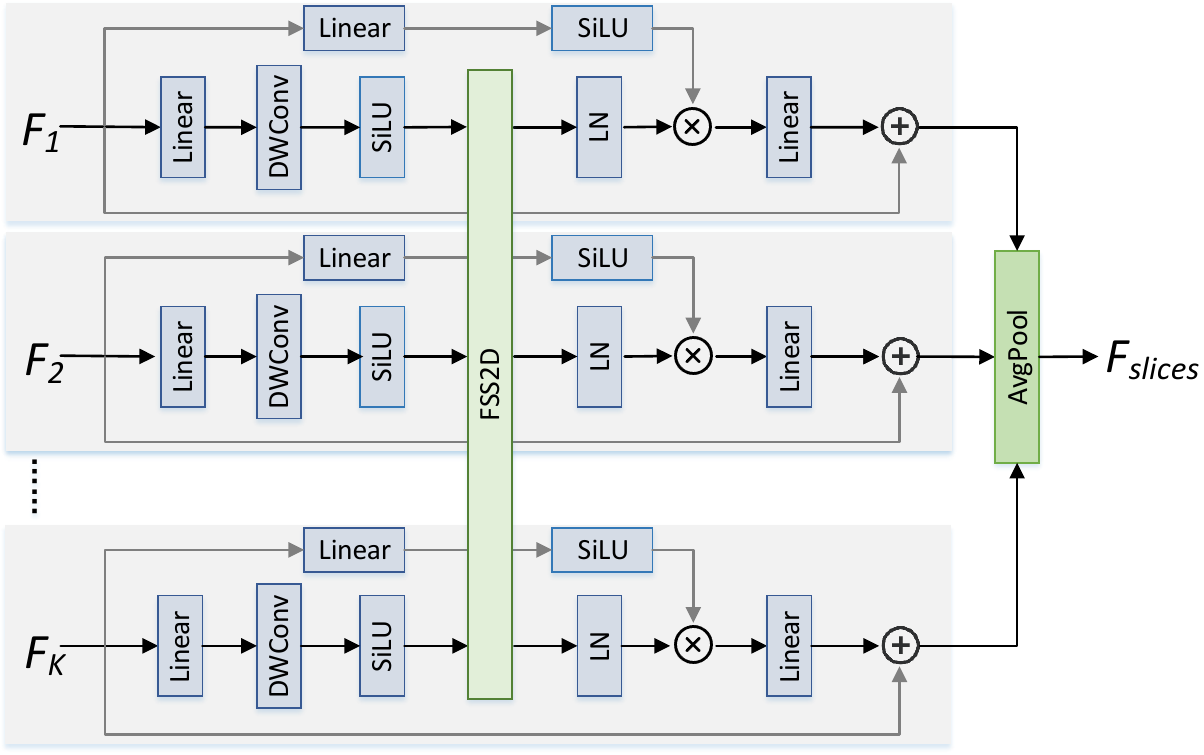}\\
\scriptsize (a) Inter-Slice Mamba\\
\includegraphics[width = 0.9\linewidth]{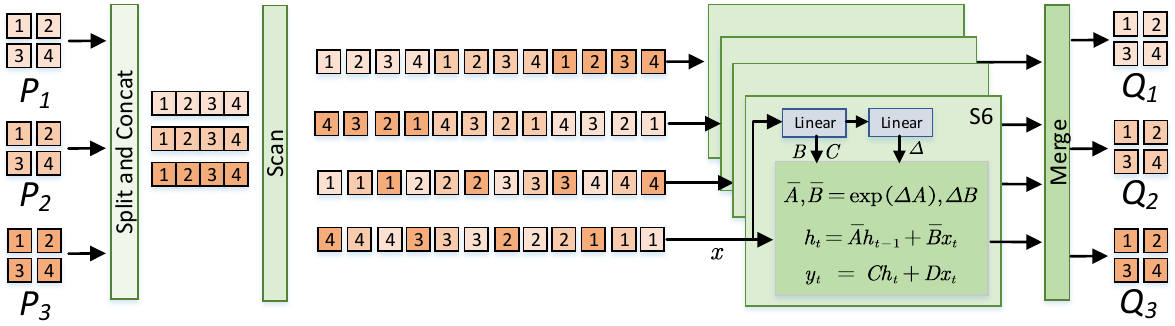}\\
\scriptsize (b) FSS2D (Taking $K$ = 3 as an example)\\
\end{tabular}
\caption{Inter-Slice Mamba and its core component FSS2D.}
\label{fig:InterSliceMamba}
\end{figure}

\textit{\textbf{Design.}}
As shown in Fig \ref{fig:InterSliceMamba} (a), the focal slice features  are respectively fed into linear projection (Linear), depth-wise convolution (DWConv), and a SiLU activation layer.
\begin{small}\begin{equation}
   P_k=SiLU(DWConv(Linear(F_k)))(k=1,\cdots,K)
\end{equation}\end{small}%
Then, Focal SS2D (FSS2D) is designed to model the long-range dependency among all the focal slices.
\begin{small}\begin{equation}
   \{Q_1,\cdots,Q_K\}=FSS2D(\{P_1,\cdots,P_K\})
\end{equation}\end{small}%
Next, a layer normalization (LN), a multiplicative branch, and a residual connection are successively followed in each slice feature, where $k=1,\cdots,K$.
\begin{small}\begin{equation}
R_k=F_k+Linear(LN(Q_k)\times SiLU(Linear(F_k)))
\end{equation}\end{small}%
Last, all the focal slice features are integrated via a concatenation and an averaged pooling operation.
\begin{small}\begin{equation}
F_{slices}=AvgPool(Concat(R_1,\cdots,R_K))
\end{equation}\end{small}
\vspace{-10mm}

Specifically,  in FSS2D illustrated in Fig \ref{fig:InterSliceMamba} (b), all the focal slice features $\{P_k\}_{k=1}^K$ are split into patches with the length $L$, and then  concatenated in the second dimension to attain an integrated feature with the size of $K\times L$. Then, the integrated feature is successively unfolded along four directions (row-wise and column-wise top-left to bottom-right, row-wise and column-wise bottom-right to top-left). Each patch sequence is fed into a separate S6 block \cite{gu2023mamba} to extract  multi-focal information along the different directions. Lastly, the four sequences are reversed to the original feature size.
By FSS2D, intra-slice and inter-slice both perform forward and backward scanning, thereby each pixel is enhanced by integrating information from all other pixels in different positions in the same slice and the same position in different slices.
The receptive fields of features are effectively enlarged to emphasize on the important information and suppress the redundancies in the focal slices.

\subsection{Feature fusion between focal slices and all-focus via Inter-Modal Mamba}
\textit{\textbf{Motivation. }}The all-focus images emphasize on the overall  appearance of the scene, while the focal slices  focus on the regions with different depths and blur the other regions. The two-modal features have been shown to be complementary \cite{liu2023lftransnet}.
Mamba constructs intra-image global receptive field by introducing data-dependent parameters ($B$, $C$, $\triangle$). In terms of two-modal features, the exchange of some parameters is beneficial for the mutual guidance, thereby proposing Slices-To-All SS2D  and the reversed  All-To-Slices  SS2D.%

\begin{figure}[!htp]
\centering
\begin{tabular}{ccc}
\includegraphics[width = 0.8\linewidth]{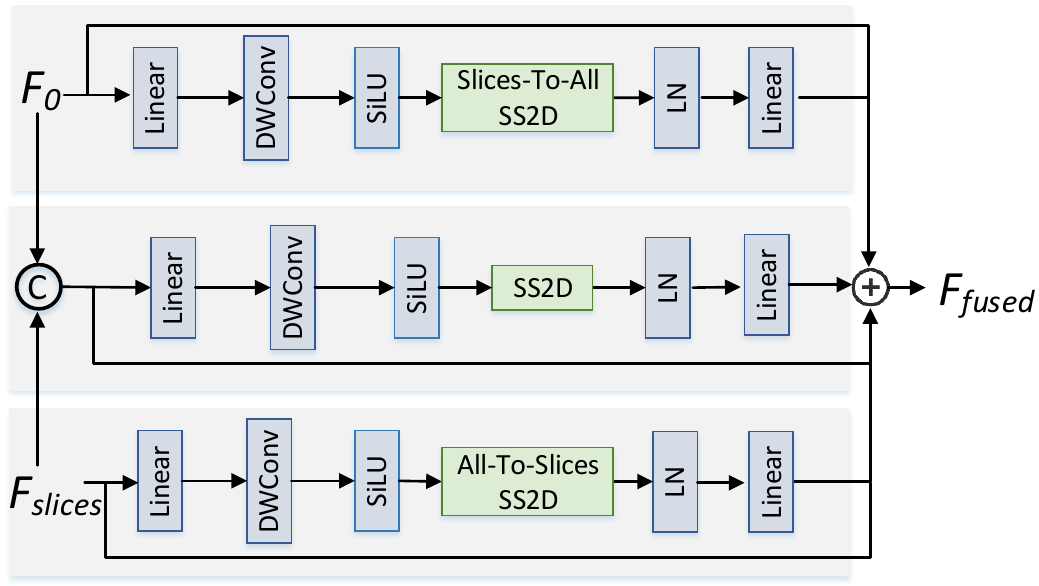}\\
\scriptsize (a) Inter-Modal Mamba\\
\includegraphics[width = 0.9\linewidth]{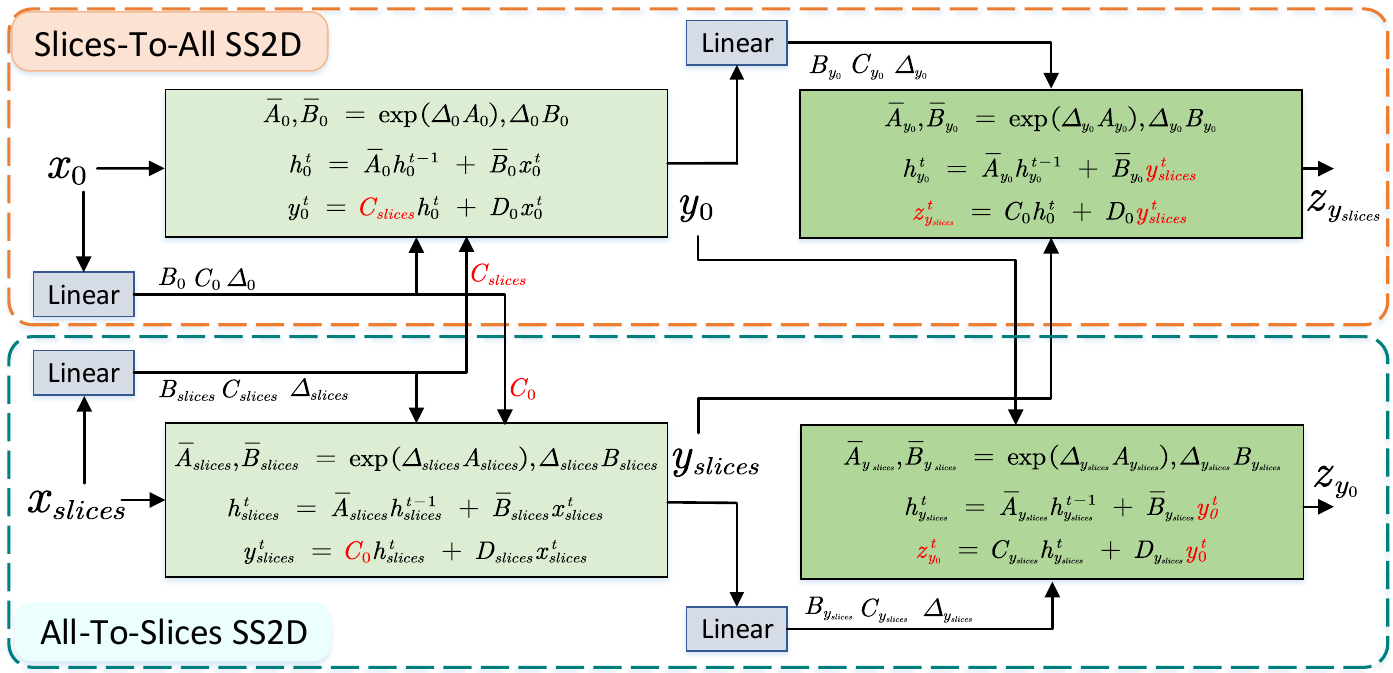}\\
\scriptsize (b) Slices-To-All SS2D and  All-To-Slices SS2D\\
\end{tabular}
\caption{Inter-Modal Mamba and its core component Slices-To-All SS2D and  All-To-Slices SS2D.}
\label{fig:CrossMamba}
\end{figure}

\textit{\textbf{Design. }}As shown in the middle stream of Fig \ref{fig:CrossMamba} (a), the all-focus feature $F_0$ and the focal slice features $F_{slices}$ are initially concatenated and convoluted to attain a basic fused feature $P$.
\begin{small}\begin{equation}
   P=Conv(Concat(F_0,F_{slices}))
\end{equation}\end{small}%

Then, the basic fused feature is successively fed into Linear, DWConv, SiLU, and an SS2D layer, followed by an LN and a Linear.
\begin{small}\begin{equation}
 \bar P=Linear(LN(SS2D(SiLU(DWConv(Linear(P))))))
\end{equation}\end{small}%

Next, in terms of all-focus $F_0$ and focal slices features $F_{slices}$ as shown in the top and bottom stream of Fig \ref{fig:CrossMamba} (a), an slices-to-all (S2A) and a reversed all-to-slices (A2S)  SS2D are applied  to achieve the mutual guidance and complements.
\begin{small}\begin{equation}
x_0=SiLU(DWConv(Linear(F_0)))
\end{equation}\end{small}%
\vspace{-5mm}
\begin{small}\begin{equation}
\bar F_0=Linear(LN(\textit{S2A}(x_0)))
\end{equation}\end{small}%
\vspace{-5mm}
\begin{small}\begin{equation}
x_{slices}=SiLU(DWConv(Linear(F_{slices})))
\end{equation}\end{small}%
\vspace{-5mm}
\begin{small}\begin{equation}
\bar F_{slices}=Linear(LN(\textit{A2S}(x_{slices})))
\end{equation}\end{small}%
%\begin{equation}
%   \hat R=Linear(LN(S2A(SiLU(DWConv(Linear(F_{slices}))))))
%\end{equation}

Last, three features are fused by addition operation after appending residual connections.
\begin{small}\begin{equation}
{F_{fused}=\bar F_0+F_0+\bar P+P+\bar F_{slices}+F_{slices}}
\end{equation}\end{small}
\vspace{-5mm}

Specifically,  in S2A and A2S SS2D, illustrated in Fig \ref{fig:InterSliceMamba} (b), two-step SS2Ds are conducted.
In the first step, the output matrixes $C_{slices}$ and $C_0$ are exchanged  to introduce the guidance of the other modality  in the output matrics.
In the second step, the output $y_0$ and $y_{slices}$ of the first step are regarded as the input. To comprehensively  interact with the information between the two modalities, $y_{slices}$ and $y_0$ are exchanged, which essentially switches the input metric, output metric, and the state transition matrix between  the two modalities so as to encourage a deeper integration.
\subsection{Decoding and supervision}
The above fused feature is last fed into SAM decoder $\mathcal{D}$ to generate the saliency map $S$.
\begin{small}\begin{equation}
    S=\mathcal{D}(F_{fused})
\end{equation}\end{small}%
\vspace{-5mm}

In fully supervised method, the weighted binary cross entropy
loss and weighted IoU loss in \cite{wei2020f3net} are adopted.
In weakly supervised method, partial cross entropy loss, local saliency coherence loss, smoothness loss in \cite{liu2023scribble} are employed.

\section{Experiments}
\subsection{Datasets}
% (100)(255) (1,462)
LF SOD dataset includes LFSD\cite{li2014saliency}, HFUT-Lytro\cite{zhang2017saliency}, and DUTLF-FS\cite{wang2019deep}. The training samples include 1,000 samples from DUTLF-FS and 100 samples from HFUT-Lytro, and the rest is used for testing. The scribble annotation is manually created by drawing both a foreground and a background scribble using the Image Labeler tool provided by MATLAB.

%Evaluation metrics adopt S-measure \cite{fan2017structure}, adaptive F-measure  \cite{achanta2009frequency}, adaptive E-measure   \cite{fan2018enhanced}, MAE \cite{perazzi2012saliency}, and Precision-recall (PR) curve \cite{borji2015salient}.
%Precision-recall (PR) curve \cite{borji2015salient} reflects the relation of precision and recall. The best model has both the best
%accuracy and the best recall.
%S-measure ($S$)  evaluates structural similarity and emphasizes structural integrity.
%F-measure ($F_\beta$) is the adaptive value presentation of the precision and recall.
%E-measure ($E_{\xi}$) captures adaptive global and local similarity.
%Mean absolute error ($M$)  measures per-pixel absolute difference.

%\subsection{Implementation Details}
%The model is trained on a NVIDIA RTX 4090 GPU.
%%Following PANet \cite{piao2021panet}, the number of focal slices in each scene is ensured to be 12 by inserting the randomly copied focal slices for coding requirements while preserving the original order.
%The input image size is $256\times256$.
%%Flipping, cropping, and rotating are used to augment the training set.
%We use AdamW optimizer with a learning rate of $5e$-$5$ and weight decay of $1e$-$4$ to train our model. The max training epoch is set to 150 and batch size is set to 2. The train process takes about 25 hours. The number of  parameters is 359M, FLOPs is  324.2G, FPS is 3.1Hz.
\subsection{Comparison with State-of-the-art Methods}
\begin{table*}[!htp]
\vspace{-5mm}

  \centering
  \fontsize{8}{10}\selectfont
  \renewcommand{\arraystretch}{1}
  \renewcommand{\tabcolsep}{1mm}
  \scriptsize
  %\captionsetup{labelformat=empty}
  \caption{Quantitative comparisons with \textbf{fully and weakly} supervised LF SOD methods on three LF SOD datasets. The best results are in bold.}
\label{tab:LFComparison}
  \scalebox{1}{
\begin{tabular}{ccccccccccccccc}
    \hline
  \multirow{2}{*}{\centering Methods} & \multirow{2}{*}{\centering Source} & \multirow{2}{*}{\centering Supervision Type}  &\multicolumn{4}{c}{\centering LFSD}&\multicolumn{4}{c}{\centering HFUT-Lytro}&\multicolumn{4}{c}{\centering DUTLF-FS}\\

  &&
  &$S\uparrow$ &$F_\beta\uparrow$ &$E_{\xi}\uparrow$&MAE$\downarrow$
  &$S\uparrow$ &$F_\beta\uparrow$ &$E_{\xi}\uparrow$&MAE$\downarrow$
  &$S\uparrow$ &$F_\beta\uparrow$ &$E_{\xi}\uparrow$&MAE$\downarrow$
   \\
\hline
MoLF\cite{zhang2019memory}&NeurIPS19&full
&.830&.819&.886&.089
&.742&.627&.785&.095
&.887&.843&.923&.052\\
DLFS\cite{piao2019deep}&IJCAI19&full
&.735&.713&.805&.149
&.741&.616&.784&.097
&.841&.801&.891&.076\\
LFNet\cite{zhang2020lfnet}&TIP20&full
&.806&.793&.870&.101
&.781&.659&.808&.076
&.882&.842&.914&.054\\
ERNet\cite{piao2020exploit}&AAAI20&full
&.835&.839&.887&.082
&.778&.705&.831&.082
&.900&.888&.942&.040\\
SA-Net\cite{zhang2021SANet}&BMVC21&full
&.841&.845&.889&.074
&.784&.736&.850&.078
&.918&.920&.954&.032\\
DLGLRG\cite{liu2021light}&ICCV21&full
&.867&.861&.898&.069
&.766&.709&.841&.071
&.928&.923&.952&.031\\
PANet\cite{piao2021panet}&TCYB21&full
&.849&.848&.893&.076
&.795&.724&.851&.074
&.908&.897&.940&.039\\
MEANet\cite{jiang2022meanet}&NEUCOM22&full
&.851&.849&.888&.077
&.796&.722&.850&.073
&.931&.920&.951&.032\\
DGENet\cite{liang2022dual}&IVC22&full
&.847&.839&.890&.075
&.771&.692&.825&.084
&.897&.881&.944&.040\\
TENet\cite{wang2023tenet}&IVC23&full
&.872&.862&.901&.063
&.838&.791&.877&.059
&.940&.940&.966&.023\\
GFRNet\cite{yuan2023guided}&ICME23&full
&.857&.859&.906&.065
&.803&.756&.852&.072
&.931&.941&.965&.026\\
LFTransNet\cite{liu2023lftransnet}&TCSVT23&full
&.905&.900&.925&.047
&.838&.775&.869&.062
&.941&.937&.962&.022\\
%FCNet\cite{zheng2024foreground}&SPIC24&full
%&.871&.881&.914&.063
%&.819&.779&.874&.067
%&.923&.930&.955&.029\\
STSA\cite{gao2023thorough}&TPAMI24&full
&.871&.872&.910&.062
&.834&.804&.880&.057
&.928&.929&.964&.027\\
FES\cite{chen2023fusion}&TMM24&full
&.888&.883&.909&.052
&.846&.800&.872&.062
&.951&.954&.971&.018\\
LFSOD-Net\cite{zheng2024spatial}&TCSVT24&full
&.889&.880&.911&.053
&.835&.784&.871&.066
&.943&.943&.963&.022\\
Ours&-&full
&\textbf{.914}&\textbf{.909}&\textbf{.929}&\textbf{.039}
&\textbf{.874}&\textbf{.821}&\textbf{.893}&\textbf{.051}
&\textbf{.957}&\textbf{.959}&\textbf{.973}&\textbf{.016}\\

         \hline
WSS\cite{wang2017learning}&CVPR17&category
&.779&.771&.837&.140
&.713&.602&.744&.122
&.771&.743&.840&.126\\
MWS\cite{zeng2019multi}&CVPR19&multiple
&.809&.785&.834&.130
&.723&.604&.732&.127
&.829&.793&.875&.104\\
SCA\cite{zhang2020weakly}&CVPR20&scribble
&.786&.782&.823&.107
&.726&.633&.779&.098
&.825&.814&.880&.075\\
%JSM\cite{li2021joint}&NeurIPS21&category
%&.745&.681&.830&.100
%&.725&.645&.773&.103
%&.768&.788&.828&.128\\
SBB\cite{liu2021weakly}&TIP21&box
&.816&.816&.858&.097
&.723&.628&.785&.099
&.832&.819&.895&.076\\
SCWS\cite{yu2021structure}&AAAI21&scribble
&.802&.794&.822&.099
&.727&.668&.788&.098
&.853&.844&.888&.063\\
%MFNet\cite{piao2021mfnet}&ICCV21&category
%&.798&.805&.860&.110
%&.747&.665&.799&.093
%&.804&.800&.873&.089\\
%HYL \cite{cong2022weakly}&TCSVT22&hybrid
%&.843&.846&.884&.082
%&.764&.710&.823&.081
%&.814&.837&.894&.084\\
WSLF \cite{liang2022weakly}&TIP22&box
&.831&.835&.880&.080
&.727&.652&.794&.097
&.889&.884&.937&.043\\
%MIRV\cite{li2023mutual}&TCSVT23&scribble
%&.877&.880&.913&.042
%&.785&.721&.835&.069
%&.849&.852&.898&.042\\
Ours&-&scribble
&\textbf{.883}&\textbf{.887}&\textbf{.902}&\textbf{.057}
&\textbf{.849}&\textbf{.790}&\textbf{.879}&\textbf{.054}
&\textbf{.918}&\textbf{.918}&\textbf{.948}&\textbf{.035}\\
    \midrule
\end{tabular}}
\end{table*}

\begin{figure}[!htp]
\begin{tabular}{ccc}
\includegraphics[width = 0.28\linewidth]{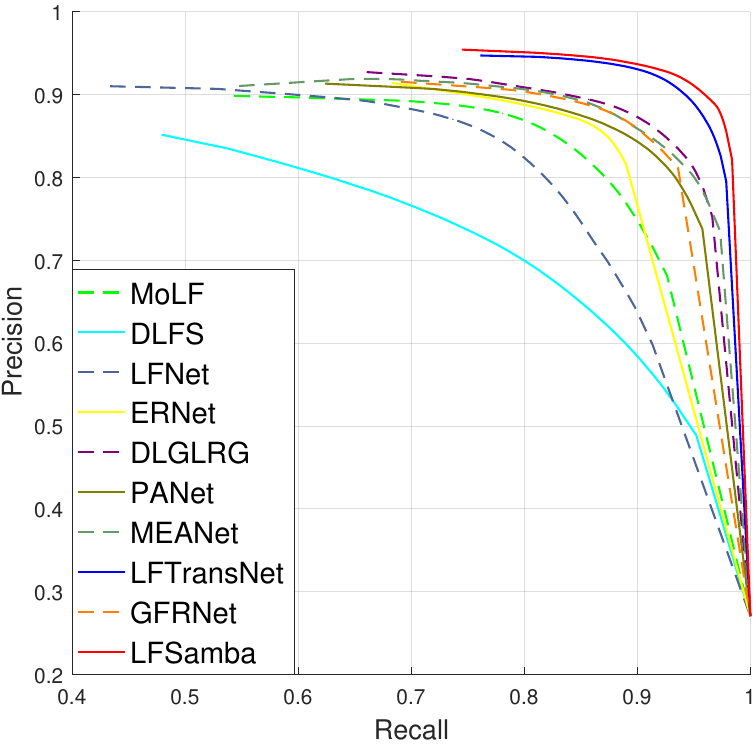}&\includegraphics[width = 0.28\linewidth]{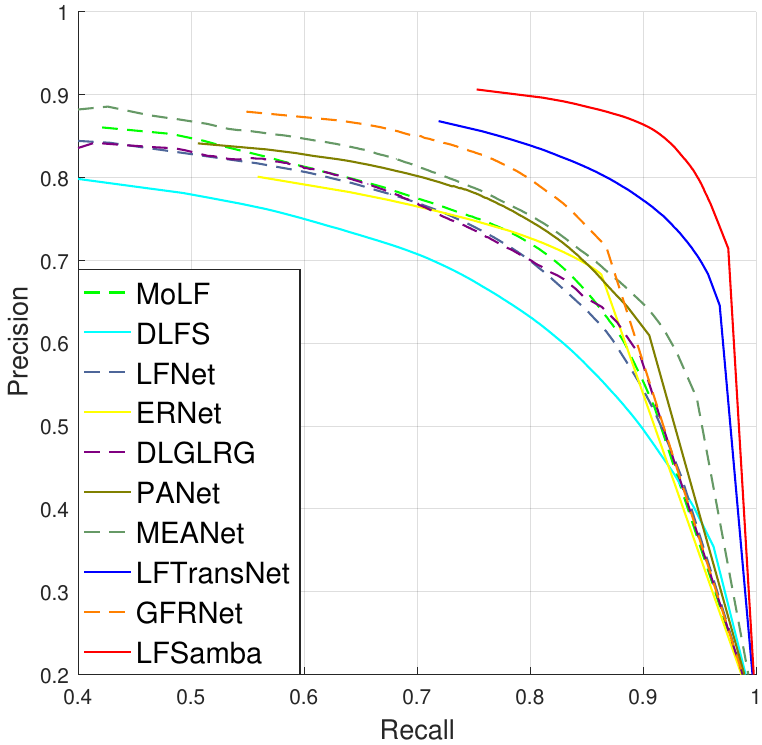}&\includegraphics[width = 0.28\linewidth]{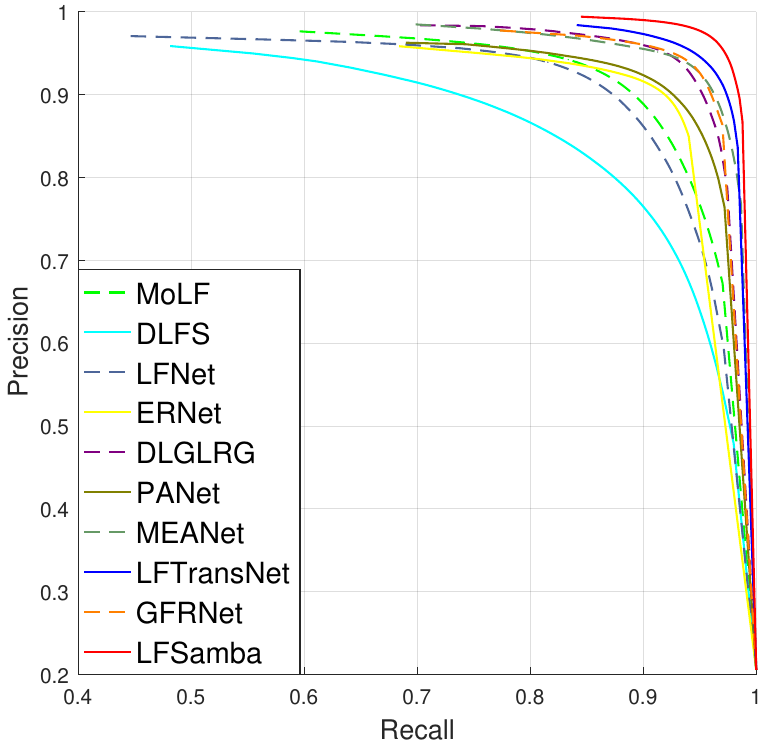}\\
\scriptsize(a) LFSD&\scriptsize(b) HFUT-Lytro&\scriptsize(c) DUTLF-FS \\
\end{tabular}
\caption{The comparison of PR curves  on three datasets. }
\label{fig:LFPRComparison}
\end{figure}

The comparison experiments of two different supervision settings are conducted on  an NVIDIA RTX 4090 GPU. Experiment settings follow LFTransNet \cite{liu2023lftransnet}. %The training time is 19 hours.
The top of Table \ref{tab:LFComparison} shows the quantitative comparisons with 15 fully-supervised methods.
%such as MoLF \cite{zhang2019memory}, DLFS \cite{piao2019deep},  LFNet \cite{zhang2020lfnet}, ERNet \cite{piao2020exploit}, SA-Net \cite{zhang2021SANet}, DLGLRG \cite{liu2021light}, PANet \cite{piao2021panet}, MEANet \cite{jiang2022meanet}, DGENet \cite{liang2022dual},
%%NoiseLF \cite{feng2022learning},
%TENet\cite{wang2023tenet},
%GFRNet\cite{yuan2023guided},
%LFTransNet\cite{liu2023lftransnet},
%%FCNet \cite{zheng2024foreground},
%STSA\cite{gao2023thorough},
%and FES \cite{chen2023fusion},
%All the compared methods are evaluated by using the same evaluation codes on the predicted saliency maps provided by the authors. We report S-measure, adaptive F-measure, adaptive E-measure, and MAE.
LFSamba outperforms the other fully supervised competitors with a large margin.
It attributes to a good feature extractor which adopts SAM with finetuned adapters, an effective focal slice feature integration method which applied Inter-Slice Mamba, an optimal multi-modal fusion method which employs Inter-Modal Mamba. Moreover, the PR curves in Fig. \ref{fig:LFPRComparison} further confirm the strength of LFSamba, as its curve approaches the upper right corner.

The bottom of Table \ref{tab:LFComparison} shows the quantitative comparisons with 6 weakly supervised methods. Our method uses sparse scribble annotation. Compared with the other weakly supervised methods, our method maintains an outstanding level. The average reduction  in MAE values across the three datasets reaches 31\%. %Moreover, our weakly supervised performance reaches a moderate level compared to fully supervised methods.%our weakly supervised performance achieves a medium level in fully supervised ones.

\subsection{Ablation studies}

\begin{figure}[!htp]

  \centering
  \includegraphics[width=0.95\linewidth]{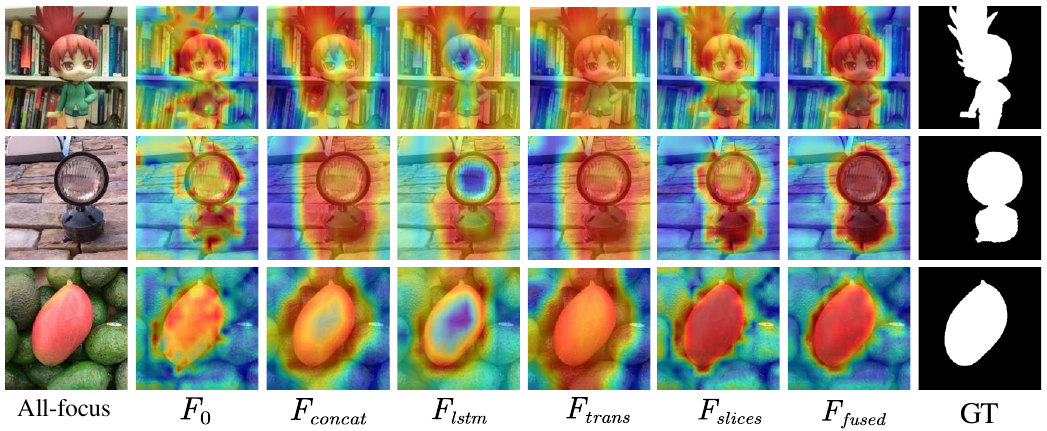}
  \caption{Feature Maps Visualization.
  \label{fig:FeatureMapCom}}
\end{figure}

Table \ref{tab:ModuleAblation} shows the contributions of all the components. SAM endowed with adapters plays an important role in improving the performance. The evaluation metrics in the first row have outperformed most existing methods.
Meanwhile, Inter-Slice Mamba and Inter-Modal Mamba work together to effectively pulls up the indicators, achieving state-of-the-art performance.
Furthermore, to verify the roles of Mambas, the result using Concatenation, ConvLSTM, and Transformer are illustrated in the last three rows. The average reduction in MAE values across the three datasets reaches 14\%, which validating the effect of Mamba.
Table \ref{tab:CostAblation} compares the computation cost. ConvLSTM is a high-cost method, while Concatenation is the least expensive. Compared to the Transformer, Mamba exhibits lower time complexity.
Fig \ref{fig:FeatureMapCom} gives the feature map visualization.
Compared $F_{concat}$, $F_{lstm}$, and $F_{trans}$ with $F_{slices}$,  the advantages of Inter-Slice Mamba in fusing focal slices are clearly evident from the object outlines. The complementary all-focus feature $F_0$ and focal slices feature $F_{slices}$ are combined through Inter-Modal Mamba to create  the optimal fused feature $F_{fused}$.
\begin{table}[!htp]
%\vspace{-5mm}
  \centering
  \fontsize{8}{10}\selectfont
  \renewcommand{\arraystretch}{1}
  \renewcommand{\tabcolsep}{1mm}
  \scriptsize
  %\captionsetup{labelformat=empty}
  \caption{Ablation study about Adapter$\diamondsuit$, Inter-Slice Mamba$\spadesuit$, and Inter-Modal Mamba$\clubsuit$  based on Baseline$\heartsuit$. The last three rows  show the results of integrating focal slices by Concatenation$\Box$, ConvLSTM$\triangle$, and Transformer$\circ$, respectively. }
\label{tab:ModuleAblation}
  \scalebox{0.85}{
\begin{tabular}{ccccccccccccc}
    \hline
  \multirow{2}{*}{\centering Variant}  &\multicolumn{4}{c}{\centering LFSD}&\multicolumn{4}{c}{\centering HFUT-Lytro}&\multicolumn{4}{c}{\centering DUTLF-FS}\\

  %&Baseline&Adapter&Inter-Slice Mamba&Inter-Modal Mamba&Addition&ConvLSTM

  &$S\uparrow$ &$F_\beta\uparrow$ &$E_{\xi}\uparrow$&MAE$\downarrow$
  &$S\uparrow$ &$F_\beta\uparrow$ &$E_{\xi}\uparrow$&MAE$\downarrow$
  &$S\uparrow$ &$F_\beta\uparrow$ &$E_{\xi}\uparrow$&MAE$\downarrow$
   \\
     \hline
$\heartsuit$
&.896&.892&.923&.050
&.817&.760&.858&.070
&.940&.942&.966&.023\\
$\heartsuit$$\diamondsuit$
&.903&.898&.927&.044
&.859&.812&.888&.059
&.948&.945&.968&.019\\
%单FSSM
$\heartsuit$$\diamondsuit$$\spadesuit$
&.911&.905&.924&.042
&.867&.807&.879&.055
&.956&.954&.968&.017\\
%单focalSS
$\heartsuit$$\diamondsuit$$\clubsuit$
&.904&.904&.920&.045
&.864&.799&.883&.053
&.954&.956&.971&.018\\
$\heartsuit$$\diamondsuit$$\spadesuit$$\clubsuit$
&\textbf{.914}&\textbf{.909}&\textbf{.929}&\textbf{.039}
&\textbf{.874}&\textbf{.821}&\textbf{.893}&\textbf{.051}
&\textbf{.957}&\textbf{.959}&\textbf{.973}&\textbf{.016}\\
 \hline
$\heartsuit$$\diamondsuit$$\Box$
&.901&.889&.925&.045
&.857&.811&.883&.057
&.953&.951&.968&.019\\
$\heartsuit$$\diamondsuit$$\triangle$
&.903&.896&.924&.046
&.860&.800&.880&.058
&.953&.950&.965&.019\\
$\heartsuit$$\diamondsuit$$\circ$
&.899&.895&.921&.045
&.867&.807&.884&.054
&.949&.940&.967&.020\\
    \hline
\end{tabular}}
%\vspace{-5mm}
\end{table}

\begin{table}[!htp]
  \centering
  \fontsize{8}{10}\selectfont
  \renewcommand{\arraystretch}{1}
  \renewcommand{\tabcolsep}{1mm}
  \scriptsize
  %\captionsetup{labelformat=empty}
  \caption{Computation cost comparison among Concatenation, ConvLSTM, and Transformer, and our Mamba. }
\label{tab:CostAblation}
  \scalebox{1}{
\begin{tabular}{ccccc}
    \hline
  Variant &Concatenation&ConvLSTM&Transformer&Our Mamba\\
    \hline
Params (M)$\downarrow$&328&387&380&359\\
FLOPs (G)$\downarrow$&305&377&369&324\\
FPS (Hz)$\uparrow$&3.1&2.8&3.0&3.1\\
    \hline
\end{tabular}}
\vspace{-3mm}
\end{table}

\section{Conclusions}
A light field salient object detection model LFSamba is proposed.
It introduces SAM to extract the discriminative features, employes Mamba to model long-range dependency among focal slices and interact all-focus and multi-focal features, and last provides a scribble supervised baseline. LFSamba achieves a significant improvement in  performance.
The model requires a significant computational cost, highlighting the need for a lightweight version in the future.

\scriptsize

\end{document}